\documentclass[11pt,a4paper]{article}
\usepackage[hyperref]{emnlp-ijcnlp-2019}
\usepackage{times}
\usepackage{amsmath}
\usepackage{mathrsfs}
\usepackage{fancyhdr,graphicx,amssymb}
\usepackage{latexsym}
\usepackage{multirow}
\usepackage{multicol}

\usepackage{url}

\aclfinalcopy 

\title{Explicit Cross-lingual Pre-training for Unsupervised Machine Translation}

\author{Shuo Ren$^\dag$$^\ddag$\thanks{\ \  Contribution during internship at MSRA.}, Yu Wu $^\S$, Shujie Liu$^\S$, Ming Zhou$^\S$, Shuai Ma$^\dag$$^\ddag$ \\
$^\dag$SKLSDE Lab, Beihang University, Beijing, China \\ $^\ddag$Beijing Advanced Innovation Center for Big Data and Brain Computing, China\\
$^\S$Microsoft Research Asia, Beijing, China \\
$^\dag$\{shuoren,mashuai\}@buaa.edu.cn\  
$^{\S}$\{Wu.Yu,shujliu,mingzhou\}@microsoft.com \\
}

\date{}

\begin{document}
\maketitle
\begin{abstract}
Pre-training has proven to be effective in unsupervised machine translation due to its ability to model deep context information in cross-lingual scenarios. However, the cross-lingual information obtained from shared BPE spaces is inexplicit and limited. 
In this paper, we propose a novel cross-lingual pre-training method for unsupervised machine translation by incorporating explicit cross-lingual training signals. 
Specifically, we first calculate cross-lingual n-gram embeddings and infer an n-gram translation table from them. With those n-gram translation pairs, we propose a new pre-training model called Cross-lingual Masked Language Model (CMLM), which randomly chooses source n-grams in the input text stream and predicts their translation candidates at each time step. Experiments show that our method can incorporate beneficial cross-lingual information into pre-trained models. 
Taking pre-trained CMLM models as the encoder and decoder, we significantly improve the performance of unsupervised machine translation. Our code is available at https://github.com/Imagist-Shuo/CMLM.
\end{abstract}

\section{Introduction}
Unsupervised machine translation has become an emerging research interest in recent years \cite{artetxe2017unsupervised, lample2017unsupervised, lample2018phrase, artetxe2018emnlp, marie2018unsupervised, ren2019unsupervised, lample2019crosslingual}. The common framework of unsupervised machine translation builds two initial translation models at first (i.e., source to target and target to source), and then does iterative back-translation \cite{sennrich2016improving, zhang2018joint} with the two models using pseudo data generated by each other. The initialization process is crucial to the final translation performance as pointed in \newcite{lample2018phrase}, \newcite{artetxe2018emnlp} and \newcite{ren2019unsupervised}. 

Previous approaches benefit mostly from cross-lingual n-gram embeddings, but recent work proves that cross-lingual language model pre-training could be a more effective way to build initial unsupervised machine translation models \cite{lample2019crosslingual}. 
However, in their method, the cross-lingual information is mostly obtained from shared Byte Piece Encoding (BPE) \cite{sennrich2016neural} spaces during pre-training, which is inexplicit and limited. Firstly, although the same BPE pieces from different languages may share the same semantic space, the semantic information of n-grams or sentences in different languages may not be shared properly. However, cross-lingual information based on n-gram level is crucial to model the initialization of unsupervised machine translation \cite{lample2018phrase,artetxe2018emnlp}, which is absent in the current pre-training method. Secondly, BPE sharing is limited to languages that share much of their alphabet. For language pairs that are not the case, the above pre-training method may not provide much useful cross-lingual information.

\begin{figure*}[!htb]
\centering
\includegraphics[width=0.8\linewidth]{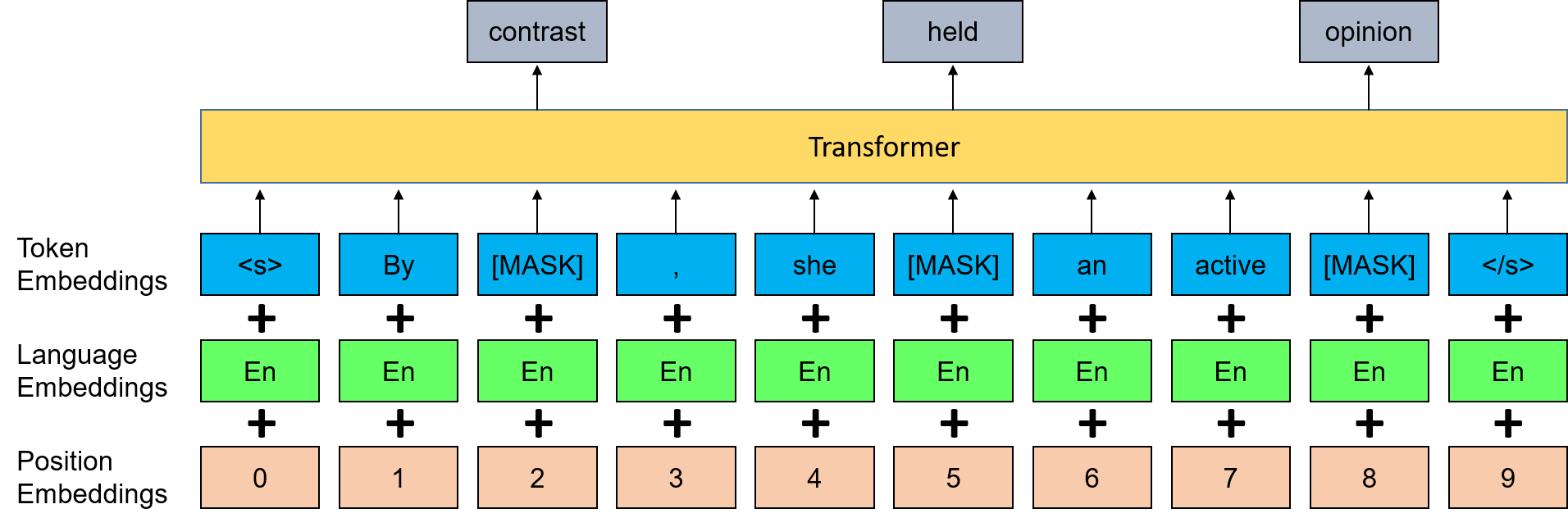}
\caption{Masked Language Model (MLM) for BERT training. For a given sentence, this task is to predict randomly masked tokens, i.e., ``contrast", ``held" and ``opinion". In practice, it is implemented based on BPE.}
\label{fig:MLM}
\end{figure*}

In this paper, by incorporating explicit cross-lingual training signals, we propose a novel cross-lingual pre-training method based on BERT \cite{devlin2018bert} for unsupervised machine translation. Our method starts from unsupervised cross-lingual n-gram embeddings, from which we infer n-gram translation pairs. Then, we propose a new pre-training objective called Cross-lingual Masked Language Model (CMLM), which masks the input n-grams randomly and predicts their corresponding n-gram translation candidates inferred above. To solve the mismatch between different lengths of the masked source and predicted target n-grams, IBM models are introduced \cite{brown1993mathematics} to derive the training loss at each time step. In this way, we can guide the model with more explicit and strong cross-lingual training signals, and meanwhile, leverage the potential of BERT to model context information. We then use two pre-trained cross-lingual language models as the encoder and decoder respectively to build desired machine translation models. Our method can be iteratively performed with the n-gram translation table updated by downstream tasks. Experiments show that our method can produce better cross-lingual representations and significantly improve the performance of unsupervised machine translation. 
Our contributions are listed as follows.
\begin{itemize}
\item We propose a novel cross-lingual pre-training method to incorporate explicit cross-lingual information into pre-trained models, which significantly improves the performance of unsupervised machine translation.
\item We introduce IBM models to calculate the step-wise training loss for CMLM, which breaks the limitation that masked n-grams and predicted ones have to be the same length during BERT training. 
\item We produce strong context-aware cross-lingual representations with our pre-training method, which helps in word alignment and cross-lingual classification tasks.
\end{itemize}

\section{Background}
\subsection{BERT}
\label{sec:BERT}
BERT \cite{devlin2018bert}, short for Bidirectional Encoder Representations from Transformers, is a powerful pre-training method for natural language processing and breaks records of many NLP tasks after corresponding fine-tuning.
The core idea of BERT is pre-training a deep bidirectional Transfomer encoder \cite{vaswani2017attention} with two training tasks. The first one is Masked Language Model (MLM) referring to the \textit{Cloze} task \cite{taylor1953cloze}, which takes a straightforward approach of masking some percentage of the input tokens at random, and then predicting them with the corresponding Transformer hidden states, as shown in Figure \ref{fig:MLM}. The second one is Next Sentence Prediction, which means to predict whether two sentences are adjacent or not. This task is designed for some tasks that need modeling the relationship between two sentences such as Question Answering (QA) and Natural Language Inference (NLI). 

\begin{figure*}[!htb]
\centering
\includegraphics[width=\linewidth]{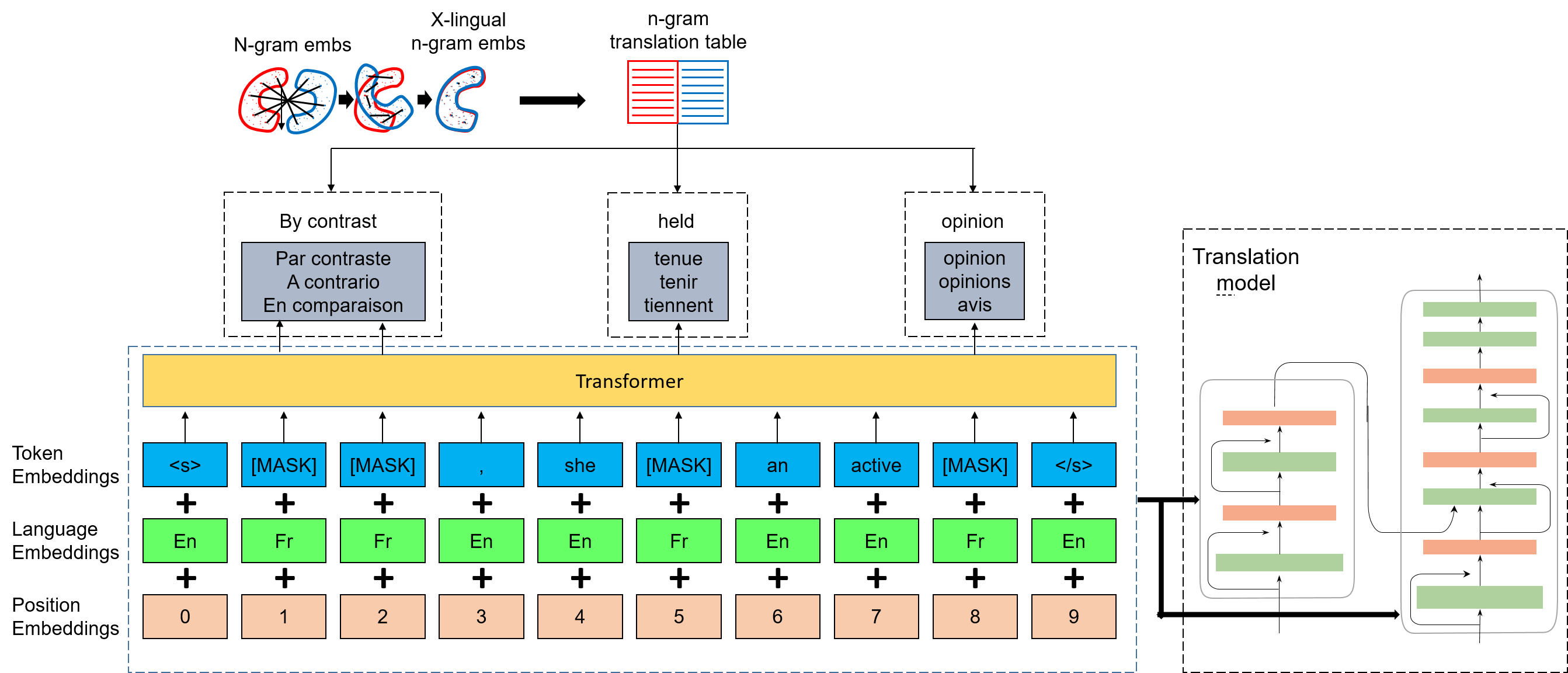}
\caption{Method overview. Our method consists of three steps. The first one is the n-gram translation table inferring. The second one is pre-training with our proposed objective Cross-lingual Masked Language Model (CMLM) which is to predict the translation candidates of randomly masked n-grams. The last step is to leverage the pre-trained cross-lingual language models as the encoder and decoder of a neural machine translation model and fine-tune the translation model iteratively.}
\label{fig:overview}
\end{figure*}

\subsection{XLM}
\label{sec:XLM}
Based on BERT, \newcite{lample2019crosslingual} propose a cross-lingual version called XLM and reach the state-of-the-art performance on some cross-lingual NLP tasks including cross-lingual classification \cite{conneau2018xnli}, machine translation, and unsupervised cross-lingual word embedding. 
The basic points of XLM are mainly two folds. The first one is to use a shared vocabulary of BPE \cite{sennrich2016neural} to provide potential cross-lingual information between two languages just as mentioned in \newcite{lample2018phrase}, in an inexplicit way though. The second point is a newly proposed training task called Translation Language Modeling (TLM), which extends MLM by concatenating parallel sentences into a single training text stream. In this way, the model can leverage the cross-lingual information provided by parallel sentences to predict the masked tokens. However, for unsupervised machine translation, TLM cannot be used due to the lack of parallel sentences. Different from them, we are motivated to give the model more explicit and strong cross-lingual information and propose a new pre-training method by (1) masking source n-grams and (2) predicting their corresponding translation candidates.

\section{Method}
\subsection{Overview}
\label{sec:overview}
Our method can be divided into three steps as shown in Figure \ref{fig:overview}. Given two languages $X$ and $Y$, we first get unsupervised cross-lingual n-gram embeddings of them, from which we infer n-gram translation tables (source-to-target and target-to-source). The n-gram translation pairs inferred in this way have proven to be instructive for initial unsupervised machine translation models \cite{artetxe2018emnlp, lample2018phrase, marie2018unsupervised, ren2019unsupervised}. Then, we pre-train cross-lingual BERT language models with our proposed Cross-lingual Masked Language Model (CMLM) objective. Specifically, we randomly choose n-grams in the monolingual sentences and predict corresponding translation candidates in the n-gram translation table inferred in the first step. 
In this way, we can guide the model with explicit and strong cross-lingual training signals. Finally, two pre-trained cross-lingual language models are used to initialize the encoder and decoder respectively, based on which, denoising auto-encoder and iterative back-translation are leveraged to fine-tune the unsupervised machine translation models. 

The above process is repeated by updating the n-gram table with the n-gram translation pairs extracted from the pseudo data generated by the translation models. In the following subsections, we will give details of each step.

\subsection{N-gram Translation Table Inferring}
\label{sec:n-gram}
Following previous work of unsupervised machine translation \cite{artetxe2018emnlp, lample2018phrase, ren2019unsupervised}, given two languages $X$ and $Y$, we build our n-gram translation tables as follows. First, we obtain monolingual n-gram embeddings using fastText \cite{bojanowski2017enriching} and then get cross-lingual n-gram embeddings using vecmap \cite{artetxe2018acl} in a fully unsupervised way. Based on that, we calculate the similarity score of n-grams $x$ and $y$ in two languages respectively with the marginal-based scoring method \cite{conneau2017word, artetxe2018margin}. Specifically, given the cross-lingual embeddings of $x$ and $y$, denoted as $e_x$ and $e_y$, the similarity score is calculated as:
\begin{equation}
\begin{aligned}	
&\textup{sim}(x,y)=\textup{margin}(\cos(e_x,e_y),\\
&\sum_{z\in\textup{NN}_{n}(x)}{\frac{\cos(e_x,e_z)}{2n}}+\sum_{z\in\textup{NN}_{n}(y)}{\frac{\cos(e_y,e_z)}{2n}})
\end{aligned}\label{eq:margin_scoring}
\end{equation}
where $\textup{margin}(a,b)$ is a marginal scoring function and $\textup{NN}_n(x)$ denotes $x$'s k-nearest neighbors in the other language. In our experiments, $n$ is 5 and $\textup{margin}(a,b)=\frac{a}{b}$. 

We take the above similarity scores as the translation probabilities between $x$ and $y$ in the n-gram table. For each top frequent n-gram in the source language, we retrieve top-$k$ n-gram translation candidates in the target language.

\subsection{Cross-lingual Masked Language Model}
\label{sec:CMLM}
In this section, we introduce our proposed method for pre-training cross-lingual language models based on BERT. Unlike the masked language model (MLM) described in Section \ref{sec:XLM} which masks several tokens in the input stream and predict those tokens themselves, we randomly select some percentage of n-grams in the input source sentence, and predict their translation candidates retrieved from Section \ref{sec:n-gram}. 
We call our proposed pre-training objective ``Cross-lingual Masked Language Model" (CMLM) as shown in Figure \ref{fig:overview}. The difficulty for BERT to predict target phrases during training is that the lengths of the translation candidates are sometimes different from the source n-grams. To deal with this problem, we turn to IBM Model 2 \cite{brown1993mathematics} to calculate the training loss at each time step. Our proposed method breaks the limitation that masked n-grams and predicted ones have to be the same length during BERT training.

Specifically, according to IBM Model 2, given a source n-gram $x_1^l$ and a target one $y_1^m$, where $l$ and $m$ are the numbers of tokens in the source and  target n-grams respectively, the translation probability from $x_1^l$ to $y_1^m$ is calculated as:
\begin{equation}
\begin{aligned}	
\textbf{Pr}(y_1^m|x_1^l)=\epsilon\prod_{j=1}^m\sum_{i=0}^l{a(i|j,l,m)p(y_j|x_i)}
\end{aligned}\label{eq:IBM2}
\end{equation}
where $\epsilon=p(m|x_1^l)$, i.e. the probability that the translation of $x_1^l$ consists of $m$ tokens; $a(j|i,l,m)$ is the probability that the $i^{th}$ source token is aligned with the $j^{th}$ target token conditioned on the lengths $l$ and $m$, while $p(y_j|x_i)$ is the translation probability from the source token $x_i$ to the target token $y_i$. Based on the IBM Model 2, the loss function of our CMLM is defined as
\begin{equation}
\begin{aligned}
\mathcal{L}_{cmlm} &=-\log{\textbf{Pr}(y_1^m|x_1^l)}\\
&= -\log\epsilon-\sum_{j=1}^m{\log{\sum_{i=0}^la(i|j,l,m)p(y_j|x_i)}}\\
\end{aligned}\label{eq:clsm_loss}
\end{equation}
The derived gradient w.r.t model parameters $\theta$ at each time step can be calculated as follows:
\begin{equation}
\begin{aligned}
&\nabla_{\theta}\mathcal{L}_{cmlm}\\
= &-\sum_{j=1}^m{\frac{a(i|j,l,m)p(y_j|x_i)}{\sum_{i=0}^l{a(i|j,l,m)p(y_j|x_i)}}\nabla_{\theta}\log{p(y_j|x_i)}}\\
\end{aligned}\label{eq:gradient}
\end{equation}

Since the target n-gram $y_1^m$ is predicted with our modified BERT, in practice, the source word $x_i$ in Eq.(\ref{eq:gradient}) is replaced with its context-sensitive embedding $\textup{C}(x_1^l)$, which is the corresponding hidden state of the top Transformer layer. The alignment probability $a(i|j,l,m)$ cannot be learned during training because of the absence of bilingual corpus. Therefore, cross-lingual BPE embeddings are leveraged to calculate the normalized $\textup{sim}(x_i,y_j)$ to approximate $a(i|j,m,l)$. $p(y_j|x_i)$ is the model prediction in Softmax outputs. For each source n-gram, all of the retrieved $k$ translation candidates are used to calculate the cross entropy loss, which are weighted with their translation probabilities in the n-gram table.

Given a language pair $X-Y$, we process both languages with the same shared BPE vocabulary using their monolingual sentences together during pre-training. Following \newcite{devlin2018bert, lample2019crosslingual}, in our CMLM objective, we randomly sample 15\% of the BPE n-grams from the text streams, and replace them by [MASK] tokens 70\% of the time. 
During pre-training, in each iteration, a batch is composed of sentences sampled from the same language, and we alternate between MLM and CMLM objectives. Different from the original MLM in BERT, in the half of the MLM time, we randomly choose some source n-grams in the input text stream, and replace them with their translation candidates to construct code-switching sentences. Our final pre-training loss is defined as 
\begin{equation}
\mathcal{L}_{pre}=\mathcal{L}_{cmlm} + \mathcal{L}_{mlm}
\label{eq:loss}
\end{equation}

\subsection{Unsupervised Machine Translation}
\label{sec:unmt}
Taking two cross-lingual language models pre-trained with the above method as the encoder and decoder, we build an initial unsupervised neural machine translation model. Then, we train the model with monolingual data until convergence via denoising auto-encoder and iterative back-translation, as described in \newcite{artetxe2017unsupervised,lample2017unsupervised,lample2018phrase}. Different from them, we step further and make another iteration with updated n-gram translation tables. Specifically, we translate the monolingual sentences with our latest translation model and run GIZA++ \cite{och2003systematic} on the generated pseudo parallel data to extract updated n-gram translation pairs, which are used to tune the encoder as Section \ref{sec:CMLM}, together with the back-translation within a multi-task learning framework. Experimental results show that running another iteration can further improve the translation performance.

It is also interesting to explore the usage of pre-trained decoders in the translation model. It seems that pre-training decoders has a smaller effect on the final performance than pre-training encoders \cite{lample2019crosslingual}, one reason for which could be that the encoder-to-decoder attention is not pre-trained. Therefore, the parameters of the decoder need to be re-adjusted substantially in the following tuning process for MT task. In our experiments, we explore some other usage of pre-trained decoders, i.e., we use the pre-trained decoder as the feature extractor and feed the outputs into a new decoder consisting of several Transformer layers with the attention to the encoder. We find this method improves the performance of some language translation directions.

\section{Experiments}
In this section, we conduct experiments to evaluate our proposed pre-training method. In Section \ref{sec:setup}, we will introduce the setup of our experiments, followed by the overall results of the final unsupervised MT models. Then, in Section \ref{sec:decoder}, we will discuss another usage of pre-trained decoders for translation models. To evaluate the cross-lingual modeling capacity of our pre-trained encoders, in Section \ref{sec:encoder}, we conduct experiments on word alignment and cross-lingual classification tasks. Finally, we do the ablation study to check the performance contribution of each component in our proposed method.

\begin{table*}[ht]
\small
\begin{center}
\begin{tabular}{l|l|cccccc}
\hline
&Method & fr2en & en2fr & de2en & en2de & ro2en & en2ro\\
\hline
\hline
\multirow{6}*{Baselines} &\cite{artetxe2017unsupervised} & 15.6 & 15.1 & - & - & - & - \\
&\cite{lample2017unsupervised} & 14.3 & 15.1 & 13.3 & 9.6 & - & - \\
&\cite{artetxe2018emnlp} & 25.9 & 26.2 & 23.1 & 18.2 & - & - \\
&\cite{lample2018phrase} & 27.7 & 28.1 & 25.2 & 20.2 & 23.9 & 25.1 \\
&\cite{ren2019unsupervised} & 28.9 & 29.5 & 26.3 & 21.7 & - & - \\
&\cite{lample2019crosslingual} & 33.3 & 33.4 & 34.3 & 26.4 & 31.8 & 33.3 \\
\hline
\hline
\multirow{2}*{Ours} & Iter 1 & 34.8 & 34.9 & 35.5 & \textbf{27.9} & 33.6 & 34.7  \\
& Iter 2 & \textbf{34.9} & \textbf{35.4} & \textbf{35.6} & 27.7 & \textbf{34.1} & \textbf{34.9} \\
\hline
\end{tabular}
\end{center}
\caption{\label{tab:overall} Comparison of the final unsupervised MT performance (BLEU). In this table, ``Iter 2" means we do the whole process with another iteration as described in Section \ref{sec:unmt}.}
\end{table*}

\subsection{Setup}
\label{sec:setup}
\subsubsection*{Data and Preprocess}
In our experiments, we consider three language pairs, English-French (en-fr), English-German (en-de) and English-Romanian (en-ro). For each language, we use all the available sentences in NewsCrawl till 2018, monolingual datasets from WMT. The NewsCrawl data are used in both pre-training and the following unsupervised NMT iteration process. Our CMLM is optimized based on the pre-trained models released by \newcite{lample2019crosslingual}\footnote{https://github.com/facebookresearch/XLM}, which are trained with Wikipedia dumps.
For fair comparison, we use \textit{newstest} 2014 as the test set for en-fr, and \textit{newstest} 2016 for en-de and en-ro.

We use Moses scripts for tokenization, and use fastBPE\footnote{https://github.com/glample/fastBPE} to split words into subword units with their released BPE codes$^1$. The number of shared BPE codes for each language pair is 60,000.

\subsubsection*{Implementation Details}
Our implementation is based on the released code of XLM$^1$ \cite{paszke2017automatic}. Specifically, we use a Transformer architecture with 1024 hidden units, 8 heads, GELU activations \cite{hendrycks2016bridging}, with a dropout rate of 0.1. The models are trained with the Adam optimizer \cite{kingma2014adam}, a linear warmup \cite{vaswani2017attention} and the learning rates varying from $10^{−4}$ to $5\times 10^{−4}$. 

For both of the MLM and CMLM objectives, we use
streams of 256 tokens and mini-batches of size 64. We use the averaged perplexity over languages as a stopping criterion for training. For machine translation, we use 6 Transformer layers, and we create mini-batches of 2000 tokens. 
\subsubsection*{Baselines}
Our method is compared with six baselines of unsupervised MT systems listed in the upper part of Table \ref{tab:overall}. The first two baselines \cite{artetxe2017unsupervised, lample2017unsupervised} use a shared encoder and different decoders for two languages with the training methods of denoising auto-encoder and iterative back-translation. The third baseline \cite{artetxe2018emnlp} is an unsupervised PBSMT model, which uses the initial PBSMT models built with language models and n-gram translation tables inferred from cross-lingual embeddings, followed with the iterative back-translation. The fourth baseline \cite{lample2018phrase} is a hybrid method of unsupervised NMT and PBSMT by combining the pseudo data generated by PBSMT models into the final iteration of NMT. The fifth baseline \cite{ren2019unsupervised} is also a hybrid method of NMT and PBSMT but different from \newcite{lample2018phrase}, they leverage PBSMT as posterior regularization in each NMT iteration. The last baseline is XLM described in Section \ref{sec:XLM}.

\subsection{Overall Results}
\label{sec:overall_results}
The overall comparison results of unsupervised machine translation are shown in Table \ref{tab:overall}. From the table, we find that our proposed method significantly outperforms previous methods on all language pairs by the average BLEU score of 1.7, and both the improvements of en2fr and ro2en are over 2 BLEU points. The results indicate that the explicit cross-lingual information incorporated by our proposed CMLM is beneficial to the unsupervised machine translation task. Notice that by doing another iteration (``Iter 2") with updated n-gram tables as described in Section \ref{sec:unmt}, we further improve the performance a bit for most translation directions with the improvements of en2fr and ro2en bigger than 0.5 BLEU point, which confirms the potential that fine-tuned machine translation models contain more beneficial cross-lingual information than the initial n-gram translation tables, which can be used to enhance the pre-trained model iteratively. 

The improvement made by \newcite{lample2019crosslingual} compared with the first five baselines shows that cross-lingual pre-training can be necessary for unsupervised MT. However, the cross-lingual information learned with this method during pre-training is mostly from the shared subword space, which is inexplicit and not strong enough. Our proposed method can give the model more explicit and strong cross-lingual training signals so that the pre-trained model contains much beneficial cross-lingual information for unsupervised machine translation. As a result, we can further improve the translation performance significantly, compared with \newcite{lample2019crosslingual} (with the significance level of p$<$0.01). 

\subsection{Usage of Pre-trained Decoder}
\label{sec:decoder}
As mentioned in Section \ref{sec:unmt}, it is interesting to explore the different usage of pre-trained decoders in the MT task. According to our intuition, directly using the pre-trained model as the decoder may not work well because parameters of the decoder need substantial adjustment due to the attention part between the encoder and the decoder. Therefore, we treat the pre-trained decoder as the feature extractor and add several Transformer layers with the encoder-to-decoder attention on top of it. We also try to fix the pre-trained decoder and just fine-tune the encoder and the added decoder part. The experiments are conducted based on ``Iter 1" with the results reported in Table \ref{tab:decoder}.

\begin{table*}[ht]
\small
\begin{center}
\begin{tabular}{l|cccccc}
\hline
Decoder type & fr2en & en2fr & de2en & en2de & ro2en & en2ro  \\
\hline
\hline
Pre-trained & \textbf{34.8} & 34.5 & \textbf{35.5} & \textbf{27.9} & 33.4 & \textbf{34.7} \\
Pre-trained + 4 TF layers & 34.2 & 33.9 & 34.9 & 26.8 & 32.5 & 33.2 \\
Pre-trained + 6 TF layers & 34.6 & \textbf{34.9} & 34.9 & 27.5 & \textbf{33.6} & 34.1 \\
Pre-trained (fix) + 4 TF layers & 26.8 & 22.0 & 23.9 & 19.2 & 24.7 & 25.9\\
Pre-trained (fix) + 6 TF layers & 28.2 & 22.4 & 24.2 & 19.7 & 25.1 & 26.2\\
\hline
\end{tabular}
\end{center}
\caption{\label{tab:decoder}Test BLEU scores with different usage of the pre-trained decoder.}
\end{table*}

From this table, we can see that directly using the pre-trained model as the decoder may be the best choice for most of the time, with the exceptions of en2fr and ro2en. By adding 6 Transformer layers on top of the original pre-trained decoder can achieve higher performance for en2fr and ro2en, but not significant. The reason could be that it is difficult to train the additional Transformer layers from scratch in the unsupervised scenario. There is also an interesting phenomenon that if we fix the pre-trained part of the decoder and only tune the added Transformer layers, the final performance will drop drastically, which indicates that the representation space of the decoder requires substantial adaptation, even though the pre-trained models already contain cross-lingual information. We think that further deep research on the decoder initialization could be a necessary and interesting topic in the future.

\subsection{Evaluation of Cross-lingual Pre-trained Encoder}
\label{sec:encoder}
\subsubsection*{Word Alignment}
To evaluate the cross-lingual modeling capacity of our pre-trained models, we first conduct experiments on the English-French (en-fr) dataset of the HLT/NAACL 2003 alignment shared task \cite{mihalcea2003evaluation}. Given two parallel sentences in English and French respectively, we feed each sentence into the pre-trained cross-lingual encoder and get its respective outputs. Then, we calculate the similarities between the outputs of the two sentences and choose target words with max similarity scores as the alignments of corresponding source words. 

We compare the context-unaware method (i.e., directly calculating the similarity scores between unsupervised cross-lingual embeddings \cite{artetxe2018acl} of source and target words), XLM \cite{lample2019crosslingual} and our proposed CMLM pre-training method in the Table \ref{tab:embed}. In this experiment, we leave out all the OOV words and those torn apart by the BPE operations.

\begin{table}[ht]
\small
\begin{center}
\begin{tabular}{l|cccc}
\hline
Method & P & R & F & AER \\
\hline
\hline
Context-unaware & 0.3860 & 0.1854 & 0.2505 & 0.6061\\
XLM & 0.5480 & 0.3178 & 0.4023 & 0.4302 \\
\hline
Ours &  \textbf{0.5898} & \textbf{0.3497} & \textbf{0.4391} & \textbf{0.4016}\\
\hline
\end{tabular}
\end{center}
\caption{\label{tab:embed}Results of word alignment tasks using different cross-lingual word embeddings. In this table, ``P" means ``precision", ``R" means “recall", ``F" means ``F-measure" and ``AER" means the ``alignment error rate".}
\end{table}

From this table, we find that,
based on BERT, both XLM and our method can model cross-lingual context information, indicating that context information can greatly enhance the cross-lingual mapping between the source and target words. By leveraging the explicit cross-lingual information in the model training, our CMLM can outperform XLM significantly. This confirms that our CMLM does better to connect the source and target representation space, with which as pre-trained models, the performance of unsupervised NMT can be improved.

\subsubsection*{Cross-lingual Classification}
We also conduct experiments on the cross-lingual classification task \cite{conneau2018xnli} using the cross-lingual language inference (XNLI) dataset \cite{conneau2018xnli}. Specifically, we add a linear classification layer on top of the first hidden state of our pre-trained model and fine-tune its parameters on the English NLI dataset. Without using any labeled data for French (fr) and Germany (de) languages, we only report the zero-shot classification results for them as shown in Table \ref{tab:xnli}. We can find that our method reaches a new record of the zero-shot cross-lingual classification task on languages of French (fr) and Germany (de), which confirms again that our CMLM works better on modeling cross-lingual information than previous methods in the unsupervised scenario.

\begin{table}[ht]
\small
\begin{center}
\begin{tabular}{l|cccc}
\hline
Method & en & fr & de \\
\hline
\hline
\cite{conneau2018xnli} & 73.7 & 67.7 & 67.7\\
\cite{devlin2018bert} & 81.4 & - & 70.5\\
\cite{lample2019crosslingual} & 83.2 & 76.5 & 74.2\\
\hline
Ours & \textbf{83.4} & \textbf{77.1} & \textbf{74.7} \\
\hline
\end{tabular}
\end{center}
\caption{\label{tab:xnli}Results of zero-shot cross-lingual classification (on XNLI test sets).}
\end{table}

\subsection{Ablation Study}
\label{sec:ablation}
In this section, we will discuss the different settings of our method. Firstly, the training loss of our pre-trained method contains two parts, i.e., CMLM and MLM, just as Eq.(\ref{eq:loss}) shows. To study the respective influences of these two parts, we remove the MLM loss from it and compare the performance on en-fr and en-de translation tasks. Since our CMLM task differs from XLM in two aspects during pre-training. The first one is that we randomly choose n-grams to mask in the input text stream rather than BPE tokens, and the second one is that we predict the translation candidates of a source n-gram rather than predicting the source n-gram itself. Although the first one has proven to be beneficial during pre-training to some NLP tasks, we want to check how much its influence is to our final translation performance. Therefore, we disable those two modifications in CMLM one by one and report the translation results. Our experiments are conducted based on ``Iter 1"  with the results in Table \ref{tab:ablation}.

\begin{table*}[ht]
\small
\begin{center}
\begin{tabular}{l|cccc}
\hline
 & fr2en & en2fr & de2en & en2de \\
\hline
CMLM + MLM & \textbf{34.8} & \textbf{34.9} & \textbf{35.5} & \textbf{27.9}\\
\hline
CMLM & 34.1 & 34.3 & 35.1 & 27.2\\
\hline
  - translation prediction & 33.7 & 33.9 & 34.8 & 26.6\\
\hline
  - - n-gram mask & 33.3 & 33.4 & 34.3 & 26.4\\
\hline
\end{tabular}
\end{center}
\caption{\label{tab:ablation}Ablation study. ``CMLM + MLM" means we use $\mathcal{L}_{pre}$ as the pre-training loss; ``CMLM" means we only use $\mathcal{L}_{cmlm}$ as the pre-training loss; ``- translation prediction" means we predict the masked n-grams themselves rather than their translation candidates during pre-training; ``- - n-gram mask" means we randomly mask BPE tokens rather than n-grams based on ``- translation prediction" during pre-training, which degrades our method to XLM.}
\end{table*}

From Table \ref{tab:ablation}, we can find that the combination of CMLM and MLM can improve the translation performance by about 0.6 to 0.7 BLEU compared with any one only. 
This confirms the monolingual context modeling capacity of the MLM, which is quite useful for unsupervised machine translation. By combining CMLM and MLM, we can enforce our model to learn both monolingual and cross-lingual information during pre-training. Besides, we find the two modifications(translation prediction and n-gram mask) made by CMLM have nearly equal contributions to the translation performance, except for en2de, where the ``n-gram mask" has little influence.

\section{Related Work}
Unsupervised machine translation dates back to \newcite{klementiev2012toward, nuhn2012deciphering}, but becomes a hot research topic in recent years. As the pioneering methods, \newcite{artetxe2017unsupervised,lample2017unsupervised,yang2018unsupervised} are mainly the modifications of the encoder-decoder structure. The core idea is to constrain outputs of encoders of two languages into a same latent space with a weight sharing mechanism such as using a shared encoder. Denoising auto-encoder \cite{vincent2010stacked} and adversarial training methods are also leveraged. Besides, they apply iterative back-translation to generated pseudo data for cross-lingual training. In addition to NMT methods for unsupervised machine translation, some following work shows that SMT methods and the hybrid of NMT and SMT can be more effective \cite{artetxe2018emnlp, lample2018phrase, marie2018unsupervised, ren2019unsupervised}. They all build unsupervised PBSMT systems, and all of their models are initialized with language models and phrase tables inferred from cross-lingual word or n-gram embeddings and then use the initial PBSMT models to do iterative back-translation. \newcite{lample2018phrase} also build a hybrid system by combining the best pseudo data that SMT models generate into the training of the NMT model while \newcite{ren2019unsupervised} alternately train SMT and NMT models with the framework of posterior regularization. 

More recently, \newcite{lample2019crosslingual} reach new state-of-the-art performance on unsupervised en-fr and en-de translation tasks. They propose a cross-lingual language model pre-training method based on BERT \cite{devlin2018bert}, and then treat two cross-lingual language models as the encoder and decoder to finish the translation. Leveraging much more monolingual data from Wikipedia, their work shows a big potential of pre-training for unsupervised machine translation. However, the cross-lingual information is obtained mostly from the shared BPE space during their pre-training method, which is inexplicit and limited. Therefore, we figure out a new pre-training method that gives the model more explicit and stronger cross-lingual information.

In the recent work of \newcite{song2019mass}, they also mask several consecutive tokens in the source sentence, but jointly pre-train the encoder and decoder by making the decoder to predict those masked tokens in both the source and target sides. Their method is a good case of pre-training for seq-to-seq tasks but the cross-lingual information incorporated with their method is from BPE sharing, which is also implicit. Our proposed method can be combined with their method within a multitask framework, which could be done in the future.

\section{Conclusion}
In this paper, we propose a novel cross-lingual pre-training method for unsupervised machine translation. In our method, we leverage Cross-lingual Masked Language Model (CMLM) to incorporate explicit and strong cross-lingual information into pre-trained models. Experimental results on en-fr, en-de, and en-ro language pairs demonstrate the effectiveness of our proposed method.

In the future, we may apply our pre-training method to other language pairs and delve into the performance of the pre-trained encoders on other NLP tasks, such as Name Entity Recognition.

\section*{Acknowledgments}
This work is supported in part by National Key R\&D Program of China {\small 2018YFB1700403}, and NSFC {\small U1636210\&61421003}.

\bibliography{emnlp-ijcnlp-2019}
\bibliographystyle{acl_natbib}

\end{document}